\pdfoutput=1

\documentclass[11pt]{article}
\usepackage{emnlp2021}
\usepackage{times}
\usepackage{latexsym}

\usepackage{pdfpages}
\usepackage{graphicx}
\usepackage{amsmath}
\usepackage{enumitem}
  \setlist{nosep}  
  \setlist[description]{labelindent=\parindent}  
\usepackage{wrapfig}
\usepackage[T1]{fontenc}

\usepackage[utf8]{inputenc}

\usepackage{microtype}
\usepackage{listings}
\usepackage{adjustbox}

\title{JaMIE: A Pipeline Japanese Medical Information Extraction System}

\author
{Fei Cheng$^{1*}$\hspace{1em}
Shuntaro Yada$^{2*}$\hspace{1em}
Ribeka Tanaka$^{1}$\thanks{$^*$ Equal contributions}\hspace{1em}\\
{\bf Eiji Aramaki$^2$\hspace{1em}}
{\bf Sadao Kurohashi$^1$\hspace{1em}}\\
$^1$Kyoto University, Kyoto, Japan\\
$^2$Nara Institute of Science and Technology, Kyoto, Japan\\
\texttt{\{feicheng, ribeka.tanaka, kuro\}@i.kyoto-u.ac.jp},\\
\texttt{\{s-yada,aramaki\}@is.naist.jp}
}

\date{}

\begin{document}
\maketitle
\begin{abstract}
We present an open-access natural language processing toolkit for Japanese medical information extraction. We first propose a novel relation annotation schema for investigating the medical and temporal relations between medical entities in Japanese medical reports. We experiment with the practical annotation scenarios by separately annotating two different types of reports. 
We design a pipeline system with three components for recognizing medical entities, classifying entity modalities, and extracting relations. 
The empirical results show accurate analyzing performance and suggest the satisfactory annotation quality, the effective annotation strategy for targeting report types, and the superiority of the latest contextual embedding models.
\end{abstract}

\section{Introduction}

Electronic medical record systems have been widely adopted in the hospitals. 
In the past decade, research efforts have been devoted to automated Information Extraction (IE) from raw medical reports. This approach should be able to liberate users from the burden of reading and understanding large volumes of records manually. While substantial progress has been made already in medical IE, it still suffers from the following limitations. 

\textbf{First}, languages are the natural boundaries to hinder the existing research from being reused across languages. The development of the English corpora and approaches can less reflect the progress in other languages. \newcite{ntcir10,ntcir11,nticr12} present a series of Japanese clinical IE shared tasks. However, more semantic-aware tasks such as medical relation extraction~\cite{uzuner20112010} and temporal relation extraction~\cite{Bethard2017} are still undeveloped. \textbf{Second}, most existing medical IE datasets focus on general report content such as discharge summary, instead of more specific report types and diseases. Such settings potentially sacrifice the accuracy for analyzing specific report types, such as radiography interpretation reports.

In this work,  we first propose a novel relation annotation scheme for investigating the medical and temporal relations in Japanese medical reports. 
Then, we intend to explore the correlation between the annotation efforts on specific report types and their analyzing accuracy, which is especially in demand for practical medical applications.
Therefore, we target the comparison of analyzing two report types involved with the diseases of high death rates: (1) specific radiography interpretation reports of lung cancer (LC), (2) medical history reports (containing multiple types of reports relevant to a patient) of idiopathic pulmonary fibrosis (IPF). The relation annotation is based on the existing entities presented by \newcite{yada-etal-2020-towards}, which annotated the medical entities (e.g. \textit{disease}, \textit{anatomical}) and their modality information (e.g. \textit{positive}, \textit{suspicious}) in Japanese medical reports.

While rich English NLP tools for medical IE have been developed such as cTAKES~\cite{savova2010mayo} and MetaMap~\cite{aronson2010overview}, there are few Japanese tools available until MedEx/J~\cite{aramaki2018medex}. MedEx/J extracts only diseases and their negation information. In this paper, we present \textsf{JaMIE}: a pipeline \textbf{Ja}panese \textbf{M}edical \textbf{IE} system, which can extract a wider range of medical information including medical entities, entity modalities, and relations from raw medical reports.

\begin{table*}[ht]
\small
\begin{center}
\begin{adjustbox}{max width=\textwidth}
\begin{tabular}{p{0.1\linewidth}|p{0.2\linewidth}|p{0.7\linewidth}}
\hline \bf Category & \bf Relation Type & \bf Example \\ \hline
Medical& $\mathrm{change}(C, A)$ & The \texttt{<A>}intrahepatic bile ducts\texttt{</A>} are \texttt{<C>}dilated\texttt{</C>}. \\
 & $\mathrm{compare}(C, \mathit{TIMEX3})$ & \texttt{<C>}Not much has changed\texttt{</C>} since \texttt{<TIMEX3>}September 2003\texttt{</TIMEX3>}. \\
 & $\mathrm{feature}(F, D)$ & No \texttt{<F>}pathologically significant\texttt{</F>} \texttt{<D>}lymph node enlargement\texttt{</D>}. \\ 
 & $\mathrm{region}(A, D)$ & There are no \texttt{<D>}abnormalities\texttt{</D>} in the \texttt{<A>}liver\texttt{</A>}. \\
 & $\mathrm{value}(T\textrm{-}\mathit{key}, T\textrm{-}\mathit{val})$ & \texttt{<T-key>} Smoking\texttt{</T-key>}: \texttt{<T-val>}20 cigarettes\texttt{</T-val>} \\
 \hline
Temporal & $\mathrm{on}(D, \mathit{TIMEX3})$ & On \texttt{<TIMEX3>}Sep 20XX\texttt{</TIMEX3>}, diagnosed as \texttt{<D>}podagra\texttt{</D>}. \\
 & $\mathrm{before}(\mathit{CC}, \mathit{TIMEX3})$ & She \texttt{<CC>}attended a cardiology clinic\texttt{</CC>}, during \texttt{<TIMEX3>}11--22 April\texttt{</TIMEX3>}. \\
 & $\mathrm{after}(C, \mathit{TIMEX3})$ & PSL 10mg/day had been kept since \texttt{<TIMEX3>}11 Aug\texttt{</TIMEX3>}, but it was \texttt{<C>}normalized\texttt{</C>}. \\
 & $\mathrm{start}(M\textrm{-}\mathit{key}, \mathit{TIMEX3})$ & \texttt{<M-key>}Equa\texttt{</M-key>} started at \texttt{<TIMEX3>}23 April\texttt{</TIMEX3>} \\
 & $\mathrm{finish}(R, \mathit{TIMEX3})$ & On \texttt{<TIMEX3>}17 Nov\texttt{</TIMEX3>}, quitting \texttt{<R>}HOT\texttt{</R>}. \\
\hline
\end{tabular}
\end{adjustbox}
\end{center}
\caption{\label{tab:examples} The example of each relation type. }
\end{table*}

In summary, we achieves three-fold contributions as following:
\begin{itemize}
\setlength{\itemsep}{0pt}
\item We present a novel annotation schema for both medical and temporal relations in Japanese medical reports.
\item We manually annotate the relations for two types of reports and empirically analyze their performance and desired annotation amount.
\item We release an open-access toolkit \textsf{JaMIE} for automatically and accurately annotating medical entities ($F1$:95.65/85.49), entity modalities ($F1$:94.10/78.06), relations ($F1$:86.53/71.04) for two report types.
\end{itemize}
Although the annotated corpus is not possible to be opened due to the increase of anonymization level, the system code and trained models are released.~\footnote{https://github.com/racerandom/JaMIE/tree/demo}

\section{Japanese Medical IE Annotation}


\subsection{Entity and Modality Annotation}

We adopt the following medical entity types defined in \newcite{yada-etal-2020-towards}: 
Diseases and symptoms \texttt{<D>}, Anatomical entities \texttt{<A>}, Features and measurements \texttt{<F>}, Change \texttt{<C>}, Time  \texttt{<TIMEX3>}, Test \texttt{<T-test/key/val>}, Medicine \texttt{<M-key/val>}, Remedy \texttt{<R>},  Clinical Context \texttt{<CC>}. The complete entity and modality definition refers to the original paper.

\subsection{Relation Annotation}

On the top of the entity and modality annotation above, we designed relation types between two entities.
They can be categorized into \textit{medical relations} and \textit{temporal relations}. The example of each relation type is presented in Table~\ref{tab:examples}.

\subsubsection{Medical Relations}

A $\mathrm{relation}(X, Y)$ denotes an entity of \texttt{<X>} type has a $\mathrm{relation}$ type toward another entity of the type \texttt{<Y>}, in which \texttt{<X>} and \texttt{<Y>} can be any entity type defined above (including the case that \texttt{<X>} is the same type as \texttt{<Y>}). 

\begin{description}
    \item[change:] A \texttt{<C>} entity changes the status of another entity, the type of which can be \texttt{<D>}, \texttt{<A>}, and \texttt{<T/M-key>}. A \texttt{<C>} is often presented as `dilate’, `shrink’, `appear’, etc.
    
    \item[compare:] A \texttt{<C>} entity's change is compared to a certain point \texttt{<Y>}, typically \texttt{<TIMEX3>}. 
    
    \item[feature:] A \texttt{<F>} entity describes a certain entity \texttt{<Y>}. A \texttt{<F>} is often presented as `significant', `mild', the size (of a tumor), etc.
    
    \item[region:] An entity of an object includes or contains another object entity (often \texttt{<D>} or \texttt{<A>}). 
    
    \item[value:] The correspondence relation between \texttt{<T/M-key>} and \texttt{<T/M-val>}. In a rare case, however, other entities of the type \texttt{<TIMEX3>} and \texttt{<D>} may correspond to a value of a \texttt{<X-key>} entity. 
    
\end{description}

\begin{figure*}[t]
\center
\includegraphics[width=0.95\linewidth]{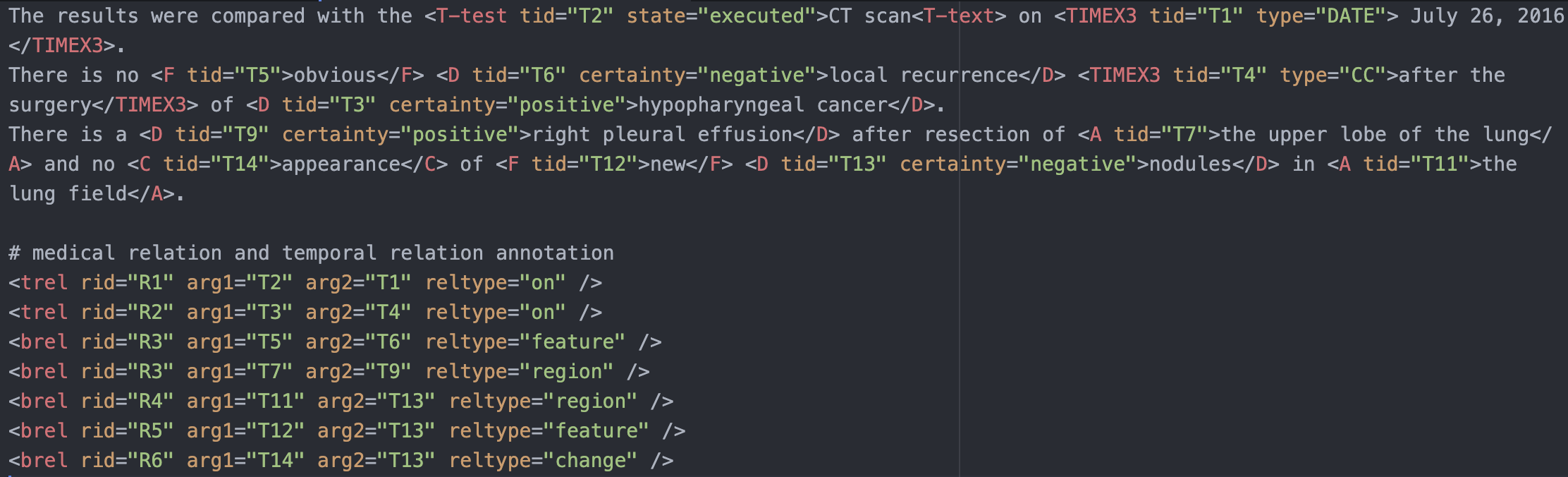}
\caption{\label{fig:example} An annotated radiography interpretation report example (translated into English). To be noticed, the translation may lead to unnatural annotation. For instance, 'after the surgery' in the second sentence is a specific temporal expression often used in Japanese clinical reports, while it look strange to be annotated with a time tag.}
\end{figure*}

\subsubsection{Temporal Relations}

\begin{figure}
  \centering
  \includegraphics[width=0.45\textwidth]{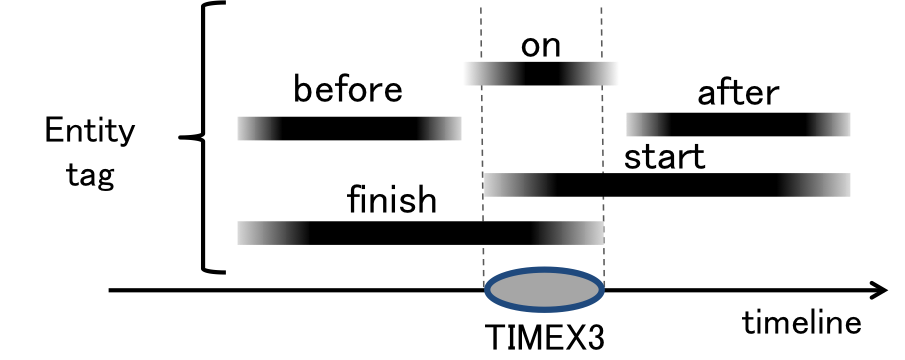}
  \caption{Visualization of temporal relations, i.e., \textit{on}, \textit{before}, \textit{after}, \textit{start}, and \textit{finish}}
  \label{fig:tlinktype}
\end{figure}

Based on an existing medical temporal-relation annotation schema, THYME \cite{Bethard2017}, we propose a simplified temporal-relation set below.
Note that any temporal relation is defined as a form $\mathrm{relation}(X, \mathit{TIMEX3})$, where the type of \texttt{<X>} can also be another \texttt{<TIMEX3>} entity.
Figure~\ref{fig:tlinktype} portrays a visualized comparison among the proposed temporal relations.

\begin{description}
    \item[on:] A \texttt{<X>} entity happens at the \textit{meantime} of a time span described by a \texttt{<TIMEX3>} entity. 
    
    \item[before:] A \texttt{<X>} entity happens \textit{before} a time span described by a \texttt{<TIMEX3>} entity. 
    
    \item[after:] A \texttt{<X>} entity happens \textit{after} a time span described by a \texttt{<TIMEX3>} entity. 
    
    \item[start:] A \texttt{<X>} entity \textit{starts} at a time span described by a \texttt{<TIMEX3>} entity. 
    
    \item[finish:] A \texttt{<X>} entity \textit{finishes} at a time span described by a \texttt{<TIMEX3>} entity. 
    
\end{description}



We show the XML-style radiography interpretation report example with the entity-level information and our relation annotation in Figure~\ref{fig:example}. The test `\texttt{<T-test>}CT scan\texttt{<T-text>}' is executed `on' the day `\texttt{<TIMEX3>}July 26, 2016\texttt{</TIMEX3>}'. A disease `\texttt{<D>}right pleural effusion\texttt{</D>}' is observed in the `region' of the anatomical entity `\texttt{<A>}the upper lobe of the lung\texttt{</A>}'. A `\texttt{<F>}new\texttt{</F>}' disease `\texttt{<D>}nodules\texttt{</D>}' is in the `region' of `\texttt{<A>}the lung field\texttt{</A>}. The `\texttt{<brel>}' and `\texttt{<trel>}' tags distinguish the medical relations and temporal relations. \textsf{JaMIE} supports this XML-style format for training models or outputting system prediction. The complete Japanese annotation guideline is available. \footnote{\url{https://sociocom.naist.jp/real-mednlp/wp-content/uploads/sites/3/2021/07/PRISM_Annotation_Guidelines.pdf}}

\subsection{Annotation}
In practice, we annotated two datasets: 1,000 radiography interpretation reports of LC and 156 medical history reports of IPF. We annotate all reports with two passes. One annotator conducted the first pass relation annotation for a report. In the second pass, the expert examined the annotation and led the final adjudication by discussing the inconsistency with the first pass annotator. This procedure is to balance the quality and cost, since it does not rely on the full expert annotation.

Table~\ref{tab:stats} shows the statistics of the relations annotation. Though the number of the medical history reports is relatively smaller, they usually contain more content per report and a wider coverage of entity types. Considering that the popular English 2010 i2b2/VA medical dataset contains 170 documents (3,106 relations) for training, our annotation scale are comparable with or even larger than it.  The results show very different relation type distribution in the two types of reports. As the medical history reports of IPF can be viewed as the mixture of several types of reports such as radiography reports, examination reports, test results, etc., they show a more balanced coverage of relation types, while the radiography interpretation reports of LC are more narrowly distributed among the disease-relevant relation types such as `region' and `feature'.

\section{System Architecture of JaMIE}

\begin{figure*}[t]
\center
\includegraphics[width=1.0\linewidth]{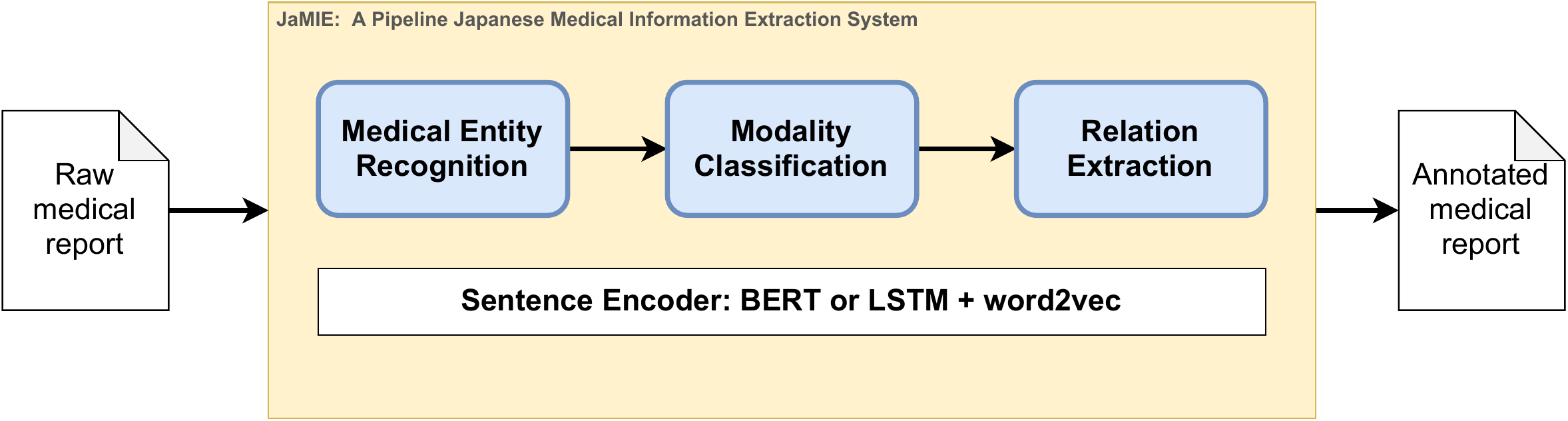}
\caption{\label{fig:system} The overview of JaMIE. }
\end{figure*}

 Figure~\ref{fig:system} shows the overview of our Japanese medical IE system with a pipeline process of three components: medical entity recognition, modality classification, and relation extraction. 

\subsection{Sentence Encoder}

Recent medical IE research~\cite{si2019enhancing,Alsentzer2019clinicalBERT,Peng2019medBERT} suggests the contextual pre-trained models such as ELMO~\cite{peters-etal-2018-deep} and BERT~\cite{devlin-etal-2019-bert} markedly outperform traditional word embedding methods (e.g., word2vec, glove, and fastText). In our pipeline system, we adopt the Japanese pre-trained BERT\footnote{https://alaginrc.nict.go.jp/nict-bert/index.html} as the sentence encoder for retrieving token embeddings.

Formally, a sentence $S = [x_0,x_1,x_2,...,x_n]$ is encoded by a contextual BERT or word embedding with bidirectional Long Short-Term Memory (LSTM)~\cite{HochSchm97} as:
\begin{equation*}
X = Encoder([x_0,x_1,x_2,...,x_n])
\end{equation*} 

\subsection{Medical Entity Recognition}

Medical entity recognition (MER) aims to predict the token spans of entities and their types from the text. 

We formulate Medical Entity Recognition as sequential tagging with the BIO (begin, inside, outside) tags. The outputs are constrained with a conditional random field (CRF)~\cite{10.5555/645530.655813} layer. For a tag sequence $y = [y_0, y_1,y_2,...,y_n]$, the probability of a sequence $y$ given $X$ is the softmax over all possible tag sequences:

\begin{equation*}
P(y|X) = \frac{e^{s(X,y)}}{\sum_{\hat{y}\in{Y}} e^{s(X,\hat{y})}}
\end{equation*}

where the score function $s(X,y)$ represents the sum of the transition scores and tag probabilities.

\begin{table*}[!ht]
\begin{center}
\begin{adjustbox}{max width=\textwidth}
\begin{tabular}{lrlr|lrlr}
\multicolumn{4}{c}{\bf 1000 Radiography Interpretation Reports (LC)} & \multicolumn{4}{c}{\bf 156 Medical History Reports (IPF)} \\
\hline \bf Med REL & \bf \#Num  & \bf Temp REL & \bf \#Num & \bf Med REL & \bf \#Num  & \bf Temp REL & \bf \#Num \\ \hline
region  & 6,794  & on  & 696 & region  & 631 & on  & 1,583 \\
change  & 689  & start  & 5  & change  & 465 & start  & 219 \\
feature  & 5,077 & finish  & 2  & feature  & 294 & finish  & 43 \\
value   & 2 & after  & 3 & value & 1,932 & after & 22 \\
compare  & 615  & before  & 1  & compare  & 229 & before  & 14 \\ 
\hline
Total & 13,884 & & & Total & 5,432 & & \\
\hline
\end{tabular}
\end{adjustbox}
\end{center}
\caption{\label{tab:stats} The statistics of the relation annotation. `\textbf{Med}' and `\textbf{Temp}' denote the medical and temporal relations.}
\end{table*}

\subsection{Modality Classification}

The modality classification (MC) component is to classify the modality types of the given entities. 
For a multi-token entity $E_i$ predicted by the MER model, we represents the entity embedding as the element sum of embeddings in the entity span. 
To enrich the context for predicting assertion, we concatenate the entity embedding with the auxiliary entity type. The $i$-step modality prediction is: 
\begin{equation*}
y_i = softmax(W[E_i; E_i^{type}] + b)
\end{equation*}

where $E_i$ denotes the $i$-th entity embedding, $E_i^{type}$ denote the entity type embedding predicted by the MER model.

\subsection{Relation Extraction}

The relation extraction (RE) component is to predict the relations and their types between two named entities. We formulate the relation extraction problem as the multiple head selection~\cite{zhang-etal-2017-dependency-parsing} of each entity in the sentence. Given each entity $E_i$ in the sentence, the model predicts whether another entity  $E_j$ is the head of this token with a relation $r_k$.  The probability of a relation is defined as: $P(E_j, r_k|E_i;\theta) = sigmoid(s(E_j, r_k, E_i))$, where $s(.)$ denotes a single full-connected layer.  An additional `N' relation presents no relation between two tokens. The final representation of an entity $E_i$ is the concatenated embeddings of the entity, entity type, and modality type.

\section{Experiments}

\subsection{Settings}
For each dataset, we conduct the patient level 5-fold cross-validation to evaluate the performance of our system. 10\% training data is split as the validation set. In each stage in the pipeline, the current component is trained with the gold inputs. The Japanese text is segmented into tokens by the morphological analyzer\footnote{MeCab~\cite{kudo-etal-2004-applying}}. We adopt PyTorch Transformers\footnote{https://github.com/huggingface/transformers} to implement the system. The following hyper-parameters are empirically chosen: fine-tuning epoch as 10, batch size as 16, AdamW with learning rate as 5e-5. The best checkpoints on the validation set are saved to produce test results. 

\begin{table*}[!ht]
\begin{center}
\begin{tabular}{l|l|ccc}
\hline \bf Report Type & \bf Encoder & \bf MER F1 & \bf MC F1 & \bf RE F1 \\ \hline
Radiography Interpretation Reports (LC) & LSTM + word2vec & 93.63 & 93.01 & 77.88 \\
 & BERT & \bf 95.65 & \bf 94.10 &  \bf 86.53 \\ 
  & \citep{yada-etal-2020-towards} & 95.30 & \bf - &  \bf - \\ \hline
Medical History Reports (IPF) & LSTM + word2vec  & 82.73 & 75.26 & 60.42 \\
 & BERT  & \bf 85.49 & \bf 78.06 & \bf 71.04 \\
\hline
\end{tabular}
\end{center}
\caption{\label{tab:pipelineresults} The main results for automatically analyzing two types of reports.}
\end{table*}

\subsection{Evaluation}
Instead of applying the usual pipeline evaluation with the gold inputs at each stage, we are more interested in the practical performance of the system and adopt the joint evaluation~\cite{zheng-etal-2017-joint} as described in the following:
\begin{itemize}[nosep]
	\item{\textbf{Medical entity recognition} identifies medical entity from raw reports. We evaluate each \textit{\{entity, entity type\}} to the reference.}
	\item{\textbf{Modality classification} classifies the modality types of the entities identified by the former stage. The evaluation is on each \textit{\{entity, entity type, modality type\}}.}
	\item{\textbf{Relation extraction} extracts the relations between the entities identified by the former stages. The evaluation is on each triplet \textit{\{entity, relation, entity2\}}.}
\end{itemize}
We measure micro-F1 of the system prediction to the gold reference in each pipeline stage.

\section{Experiment Results}
\subsection{Main Performance of JaMIE}
Table~\ref{tab:pipelineresults} shows our system performance on two types of reports:  radiography interpretation reports of LC and medical history reports of IPF.
The radiography interpretation reports' performance suggests that by concentrating annotation efforts on a specific report type the system achieves high F1 with sufficient training data. Compared to 95.30 MER F1 reported by~\cite{yada-etal-2020-towards}, our MER score outperforms their score by 0.35 with the additional CRF layer. The RE model obtains 86.53 F1 of the radiography interpretation reports and 71.04 F1 of the medical history reports. 

We offer the baseline encoder with LSTM\footnote{LSTM and Word2vec hidden size equal to 256.} upon word2vec embeddings~\cite{mikolov2013efficient} trained on Japanese Wikipedia. We observe significant drops in all three tasks, especially in the final relation extraction. In both radiography interpretation reports of LC and medical history reports of IPF, the BERT-based RE models leading `LSTM + word2vec' by approximately 10 points F1. We suggest that solving relation extraction requires long-range information between entities. BERT naturally models such long-range dependency in the self-attention mechanism, while word2vec is trained with a fixed local window and LSTM could also fall in the long sequential actions.
 
\begin{table}[!ht]
\begin{center}
\begin{tabular}{lc|lc}
\hline \bf Med REL & \bf RE F1 & \bf Temp REL & \bf RE F1 \\ \hline
region  & 84.59  & on  & 81.92 \\
change  & 76.23  & start  & 20.00 \\
feature  & 90.16 & finish  & - \\
value   &  -  & after & - \\
compare  & 80.86 & before  & - \\
\hline
\end{tabular}
\end{center}
\caption{\label{tab:rirrel} Each relation type F1 (BERT-based) in the radiography interpretation reports of LC.  }
\begin{center}
\begin{tabular}{lc|lc}
\hline \bf Med REL & \bf RE F1 & \bf Temp REL & \bf RE F1 \\ \hline
region  & 71.73  & on  & 70.48 \\
change  & 58.66  & start  & 49.33 \\
feature  & 60.54 & finish  & 12.02 \\
value   & 83.12  & after & - \\
compare  & 75.47 & before  & 11.38 \\
\hline
\end{tabular}
\end{center}
\caption{\label{tab:gmrrel} Each relation type F1 (BERT-based) in the medical history reports of IPF. }
\end{table}

While the medical history reports contain broader relation types and the data size is relatively smaller, the system still obtains satisfactory performance. In addition, we present each relation F1 in Table~\ref{tab:gmrrel}. Except for three rarely appearing relations, i.e. `finish', `after' and `before', the F1 scores on the other types are balanced and match the statistics in Table~\ref{tab:stats}. As for the radiography interpretation report results in Table~\ref{tab:rirrel}, the major relations of `region' and `feature' relations achieve high performance with 84.59 and 90.16 F1. The moderate `change', `compare' and `on' obtain satisfying 76.23 to 81.92 F1.

\subsection{Correlation between Report Types and Demanding Annotation Efforts}
\label{subsec:abla}

\begin{table}[!ht]
\begin{center}
\begin{tabular}{lc}
\hline \bf Report Type & \bf RE F1  \\ \hline
Radiography Interpretation Reports  & 86.53  \\
\hfill - with 39\% training data  & 82.33   \\ \hline
Medical history Reports & 71.04  \\
\hline
\end{tabular}
\end{center}
\caption{\label{tab:abla} The RE performance comparison between the radiography interpretation reports and medical history reports with comparable training size }
\end{table}

One question is whether concentrating annotation efforts on a specific report type can quickly obtain high accuracy to meet the requirements of the practical applications. A valid approach is to compare the RE performance of two report types with the comparable annotation efforts i.e. training data size. The medical history report of IPF contains total 5,432 relations, which is approximately 39\% of the radiography interpretation reports. We designed the experiment by reducing the train set of the radiography interpretation reports to the comparable 39\% of the origin. The results in Table~\ref{tab:abla} show that even with comparable training size, the specific radiography interpretation reports lead the performance by 11.29 points F1. 

To be clarified, the two results are still not exactly comparable due to the different relation distributions. However, the radiography interpretation reports more densely spread in the relation types such as `region' and `feature' (Table~\ref{tab:stats}), which usually means less number of report needed for achieving the similar overall accuracy compared to the medical history reports. In the scenario of demanding high accuracy for practical medical applications, the results suggest that the annotation strategy of starting from a specific type of report and gradually increasing the coverage of report types is feasible.

\section{System Application}
\subsection{User Interface}
\textsf{JaMIE} provides an easy-to-use Command-Line Interface (CLI). We design our training/testing scripts similar to the official Transformers examples, in order to be friendly to the Transformers users. We demonstrate how to train/test a relation model with the following script:
\begin{lstlisting}[language=bash, basicstyle=\small,]
  $ # Training
  $ python clinical_pipeline_rel.py \ 
  $ --pretrained_model $JAPANESE_BERT \
  $ --saved_model $MODEL_TO_SAVE \
  $ --train_file $TRAIN_FILE \
  $ --dev_file $DEV_FILE \
  $ --batch_size 16 \
  $ --do_train 
\end{lstlisting}

\begin{lstlisting}[language=bash, basicstyle=\small,]
  $ # Testing
  $ python clinical_pipeline_rel.py \
  $ --saved_model $TRAINED_MODEL \
  $ --test_file $TEST_FILE \
  $ --test_out $TEST_OUTFILE \
\end{lstlisting}

\subsection{Use Case}
In the case of annotating raw medical reports with our trained model, users need to download our trained models from the \textsf{JaMIE} GitHub beforehand. Users then execute the pipeline `test' scripts to annotate entities, modalities, and relations step by step. At each stage, the model will generate the prediction as the input of the next stage model. The prediction is presented in the same XML-style as shown in Figure~\ref{fig:example}.

Our medical IE annotation schema serves to encode a wide range of general medical information not limited to any specific disease, report types or languages. Users can manually annotate other types of medical reports by following our guideline. Users can apply the `train' scripts to train the pipeline models on their newly annotated corpus for providing automatic annotation.

\section{Conclusion}
We propose a novel annotation schema for investigating medical and temporal relations between medical entities in Japanese medial reports. We empirically compare the annotation on two types of reports: specific radiography interpretation reports of LC and medical history reports of IPF. 
The system obtains overall satisfactory performance in three tasks, supporting the valuable findings of the good annotation quality, the feasible annotation strategies for targeting report types, and the superior performance of the contextual BERT encoder. The system code and trained models on our annotation are open-access.
In the future, we plan to stick to LC and IPF, cover more specific report types involved with LC, and increase the annotation amount of medical history reports of IPF.

\bibliographystyle{acl_natbib}
\bibliography{acl2021}

\begin{thebibliography}{20}
\expandafter\ifx\csname natexlab\endcsname\relax\def\natexlab#1{#1}\fi

\bibitem[{Alsentzer et~al.(2019)Alsentzer, Murphy, Boag, Weng, Jin, Naumann,
  and McDermott}]{Alsentzer2019clinicalBERT}
Emily Alsentzer, John~R. Murphy, Willie Boag, Wei{-}Hung Weng, Di~Jin, Tristan
  Naumann, and Matthew B.~A. McDermott. 2019.
\newblock \href {http://arxiv.org/abs/1904.03323} {Publicly available clinical
  {BERT} embeddings}.
\newblock \emph{CoRR}, abs/1904.03323.

\bibitem[{Aramaki et~al.(2014)Aramaki, Morita, Kano, and Ohkuma}]{ntcir11}
Eiji Aramaki, Mizuki Morita, Yoshinobu Kano, and Tomoko Ohkuma. 2014.
\newblock Overview of the ntcir-11 mednlp-2 task.
\newblock In \emph{In Proceedings of the 11th NTCIR Workshop Meeting on
  Evaluation of Information Access Technologies}.

\bibitem[{Aramaki et~al.(2016)Aramaki, Morita, Kano, and Ohkuma}]{nticr12}
Eiji Aramaki, Mizuki Morita, Yoshinobu Kano, and Tomoko Ohkuma. 2016.
\newblock Overview of the ntcir-12 mednlpdoc task.
\newblock In \emph{In Proceedings of the 12th NTCIR Workshop Meeting on
  Evaluation of Information Access Technologies}.

\bibitem[{Aramaki et~al.(2018)Aramaki, Yano, and Wakamiya}]{aramaki2018medex}
Eiji Aramaki, Ken Yano, and Shoko Wakamiya. 2018.
\newblock Medex/j: A one-scan simple and fast nlp tool for japanese clinical
  texts.
\newblock In \emph{MEDINFO 2017: Precision Healthcare Through Informatics:
  Proceedings of the 16th World Congress on Medical and Health Informatics},
  volume 245, page 285. IOS Press.

\bibitem[{Aronson and Lang(2010)}]{aronson2010overview}
Alan~R Aronson and Fran{\c{c}}ois-Michel Lang. 2010.
\newblock An overview of metamap: historical perspective and recent advances.
\newblock \emph{Journal of the American Medical Informatics Association},
  17(3):229--236.

\bibitem[{Bethard et~al.(2017)Bethard, Savova, Palmer, and
  Pustejovsky}]{Bethard2017}
Steven Bethard, Guergana Savova, Martha Palmer, and James Pustejovsky. 2017.
\newblock \href {https://doi.org/10.18653/v1/S17-2093} {{SemEval-2017 Task 12:
  Clinical TempEval}}.
\newblock In \emph{Proceedings of the 11th International Workshop on Semantic
  Evaluation (SemEval-2017)}, pages 565--572, Stroudsburg, PA, USA. Association
  for Computational Linguistics.

\bibitem[{Devlin et~al.(2019)Devlin, Chang, Lee, and
  Toutanova}]{devlin-etal-2019-bert}
Jacob Devlin, Ming-Wei Chang, Kenton Lee, and Kristina Toutanova. 2019.
\newblock \href {https://doi.org/10.18653/v1/N19-1423} {{BERT}: Pre-training of
  deep bidirectional transformers for language understanding}.
\newblock In \emph{Proceedings of the 2019 Conference of the North {A}merican
  Chapter of the Association for Computational Linguistics: Human Language
  Technologies, Volume 1 (Long and Short Papers)}, pages 4171--4186,
  Minneapolis, Minnesota. Association for Computational Linguistics.

\bibitem[{Hochreiter and Schmidhuber(1997)}]{HochSchm97}
Sepp Hochreiter and Jürgen Schmidhuber. 1997.
\newblock Long short-term memory.
\newblock \emph{Neural Computation}, 9(8):1735--1780.

\bibitem[{Kudo et~al.(2004)Kudo, Yamamoto, and
  Matsumoto}]{kudo-etal-2004-applying}
Taku Kudo, Kaoru Yamamoto, and Yuji Matsumoto. 2004.
\newblock \href {https://www.aclweb.org/anthology/W04-3230} {Applying
  conditional random fields to {J}apanese morphological analysis}.
\newblock In \emph{Proceedings of the 2004 Conference on Empirical Methods in
  Natural Language Processing}, pages 230--237, Barcelona, Spain. Association
  for Computational Linguistics.

\bibitem[{Lafferty et~al.(2001)Lafferty, McCallum, and
  Pereira}]{10.5555/645530.655813}
John~D. Lafferty, Andrew McCallum, and Fernando C.~N. Pereira. 2001.
\newblock Conditional random fields: Probabilistic models for segmenting and
  labeling sequence data.
\newblock In \emph{Proceedings of the Eighteenth International Conference on
  Machine Learning}, ICML '01, page 282–289, San Francisco, CA, USA. Morgan
  Kaufmann Publishers Inc.

\bibitem[{Mikolov et~al.(2013)Mikolov, Chen, Corrado, and
  Dean}]{mikolov2013efficient}
Tomas Mikolov, Kai Chen, Greg Corrado, and Jeffrey Dean. 2013.
\newblock \href {http://arxiv.org/abs/1301.3781} {Efficient estimation of word
  representations in vector space}.

\bibitem[{Morita et~al.(2013)Morita, Kano, Ohkuma, Miyabe, and
  Aramaki}]{ntcir10}
Mizuki Morita, Yoshinobu Kano, Tomoko Ohkuma, Mai Miyabe, and Eiji Aramaki.
  2013.
\newblock Overview of the ntcir-10 mednlp task.
\newblock In \emph{In Proceedings of NTCIR-10}.

\bibitem[{Peng et~al.(2019)Peng, Yan, and Lu}]{Peng2019medBERT}
Yifan Peng, Shankai Yan, and Zhiyong Lu. 2019.
\newblock \href {http://arxiv.org/abs/1906.05474} {Transfer learning in
  biomedical natural language processing: An evaluation of {BERT} and elmo on
  ten benchmarking datasets}.
\newblock \emph{CoRR}, abs/1906.05474.

\bibitem[{Peters et~al.(2018)Peters, Neumann, Iyyer, Gardner, Clark, Lee, and
  Zettlemoyer}]{peters-etal-2018-deep}
Matthew Peters, Mark Neumann, Mohit Iyyer, Matt Gardner, Christopher Clark,
  Kenton Lee, and Luke Zettlemoyer. 2018.
\newblock \href {https://doi.org/10.18653/v1/N18-1202} {Deep contextualized
  word representations}.
\newblock In \emph{Proceedings of the 2018 Conference of the North {A}merican
  Chapter of the Association for Computational Linguistics: Human Language
  Technologies, Volume 1 (Long Papers)}, pages 2227--2237, New Orleans,
  Louisiana. Association for Computational Linguistics.

\bibitem[{Savova et~al.(2010)Savova, Masanz, Ogren, Zheng, Sohn,
  Kipper-Schuler, and Chute}]{savova2010mayo}
Guergana~K Savova, James~J Masanz, Philip~V Ogren, Jiaping Zheng, Sunghwan
  Sohn, Karin~C Kipper-Schuler, and Christopher~G Chute. 2010.
\newblock Mayo clinical text analysis and knowledge extraction system (ctakes):
  architecture, component evaluation and applications.
\newblock \emph{Journal of the American Medical Informatics Association},
  17(5):507--513.

\bibitem[{Si et~al.(2019)Si, Wang, Xu, and Roberts}]{si2019enhancing}
Yuqi Si, Jingqi Wang, Hua Xu, and Kirk Roberts. 2019.
\newblock Enhancing clinical concept extraction with contextual embeddings.
\newblock \emph{Journal of the American Medical Informatics Association},
  26(11):1297--1304.

\bibitem[{Uzuner et~al.(2011)Uzuner, South, Shen, and DuVall}]{uzuner20112010}
{\"O}zlem Uzuner, Brett~R South, Shuying Shen, and Scott~L DuVall. 2011.
\newblock 2010 i2b2/va challenge on concepts, assertions, and relations in
  clinical text.
\newblock \emph{Journal of the American Medical Informatics Association},
  18(5):552--556.

\bibitem[{Yada et~al.(2020)Yada, Joh, Tanaka, Cheng, Aramaki, and
  Kurohashi}]{yada-etal-2020-towards}
Shuntaro Yada, Ayami Joh, Ribeka Tanaka, Fei Cheng, Eiji Aramaki, and Sadao
  Kurohashi. 2020.
\newblock \href {https://www.aclweb.org/anthology/2020.lrec-1.561} {Towards a
  versatile medical-annotation guideline feasible without heavy medical
  knowledge: Starting from critical lung diseases}.
\newblock In \emph{Proceedings of the 12th Language Resources and Evaluation
  Conference}, pages 4565--4572, Marseille, France. European Language Resources
  Association.

\bibitem[{Zhang et~al.(2017)Zhang, Li, Lang, Xia, and
  Zhang}]{zhang-etal-2017-dependency-parsing}
Yue Zhang, Zhenghua Li, Jun Lang, Qingrong Xia, and Min Zhang. 2017.
\newblock \href {https://www.aclweb.org/anthology/I17-1006} {Dependency parsing
  with partial annotations: An empirical comparison}.
\newblock In \emph{Proceedings of the Eighth International Joint Conference on
  Natural Language Processing (Volume 1: Long Papers)}, pages 49--58, Taipei,
  Taiwan. Asian Federation of Natural Language Processing.

\bibitem[{Zheng et~al.(2017)Zheng, Wang, Bao, Hao, Zhou, and
  Xu}]{zheng-etal-2017-joint}
Suncong Zheng, Feng Wang, Hongyun Bao, Yuexing Hao, Peng Zhou, and Bo~Xu. 2017.
\newblock \href {https://doi.org/10.18653/v1/P17-1113} {Joint extraction of
  entities and relations based on a novel tagging scheme}.
\newblock In \emph{Proceedings of the 55th Annual Meeting of the Association
  for Computational Linguistics (Volume 1: Long Papers)}, pages 1227--1236,
  Vancouver, Canada. Association for Computational Linguistics.

\end{thebibliography}


\end{document}